# Land Cover Change Detection via Semantic Segmentation

Renee Su[1], Rong Chen[2]

[1]Henry M. Gunn High School, Palo Alto, CA 94306, USA
[2]Rutgers University, Piscataway, NJ 08854, USA

**Abstract:** This paper presents a change detection method that identifies land cover changes from aerial imagery, using semantic segmentation, a machine learning approach. We present a land cover classification training pipeline with Deeplab v3+, state-of-the-art semantic segmentation technology, including data preparation, model training for seven land cover types, and model exporting modules. In the land cover change detection system, the inputs are images retrieved from Google Earth at the same location but from different times. The system then predicts semantic segmentation results on these images using the trained model and calculates the land cover class percentage for each input image. We see an improvement in the accuracy of the land cover semantic segmentation model, with a mean IoU of 0.756 compared to 0.433, as reported in the DeepGlobe land cover classification challenge. The land cover change detection system that leverages the state-of-the-art semantic segmentation technology is proposed and can be used for deforestation analysis, land management, and urban planning.
**Keywords:** land cover; change detection; semantic segmentation; Google Earth; land use

## 1. Introduction

In the past year, wildfires in California have been a major source of destruction to both wildlife and human civilization. However, deforestation is not just a problem in one particular state—it is a global issue. Now, more than ever, it is important to effectively monitor these changes in order to immediately initiate future planning. One possible way to track the damage and recovery of trees is through one particular analysis of satellite data conducted by artificial intelligence, called semantic segmentation. Semantic segmentation is the labelling of each pixel in an image with a category label (Li, et al., 2018), and can be applied to satellite and aerial imagery in remote sensing field.

Previously, Mas et al. used image color segmentation, a non-machine learning approach to assess land use/cover changes (Mas, el al, 2015). Helber et al. proposed a patch-based classification method (Helber et al. 2019) to identify land cover types, where an image is cropped to 64x64 patches and each patch produces a classification label. The limitation of this method is incorrect classification in patches with multiple land cover types.

A machine learning approach to semantic segmentation generally implements the commonly used Convolutional Neural Networks (CNN). A typical CNN layout starts with an input image with three color channels, in which filters (square matrices with weighted values) are slid across each pixel on the original image, and a dot product is computed (Li, et al., 2018). This is followed by an activation function, called ReLU, which essentially enlarges targeted values while greatly decreasing other values. Since CNN's are mainly used for image classification, the images outputted after each convolutional layer can be pooled, or downsampled (pick maximum between two pixels), so as to reduce the image resolution. Downsampling layers, also called pooling layers, summarize the outputs of neighboring groups of neurons in the same kernel map (Krizhevsky, 2012). This lowers the resolution of the resultant image. Upsampling layers, on the other hand, increase the dimensions of the output image, which also increases the resolution. At the end, a fully connected layer is added to calculate the probability scores of each possible class that the image could be labelled as. Semantic segmentation differs in the sense that each individual pixel needs to be labelled, rather than the whole image. This means that downsampling is not optimal, as trying to classify individual pixels without regard for surrounding pixels is computationally expensive. A basic proposal for semantic segmentation is to downsample to a certain point, and then perform upsampling, which intends to restore the resulting image to its original resolution. One can use a transpose convolution (also called deconvolutions and fractionally strided convolutions) to increase the dimensions after each operation (Li, et al., 2018).

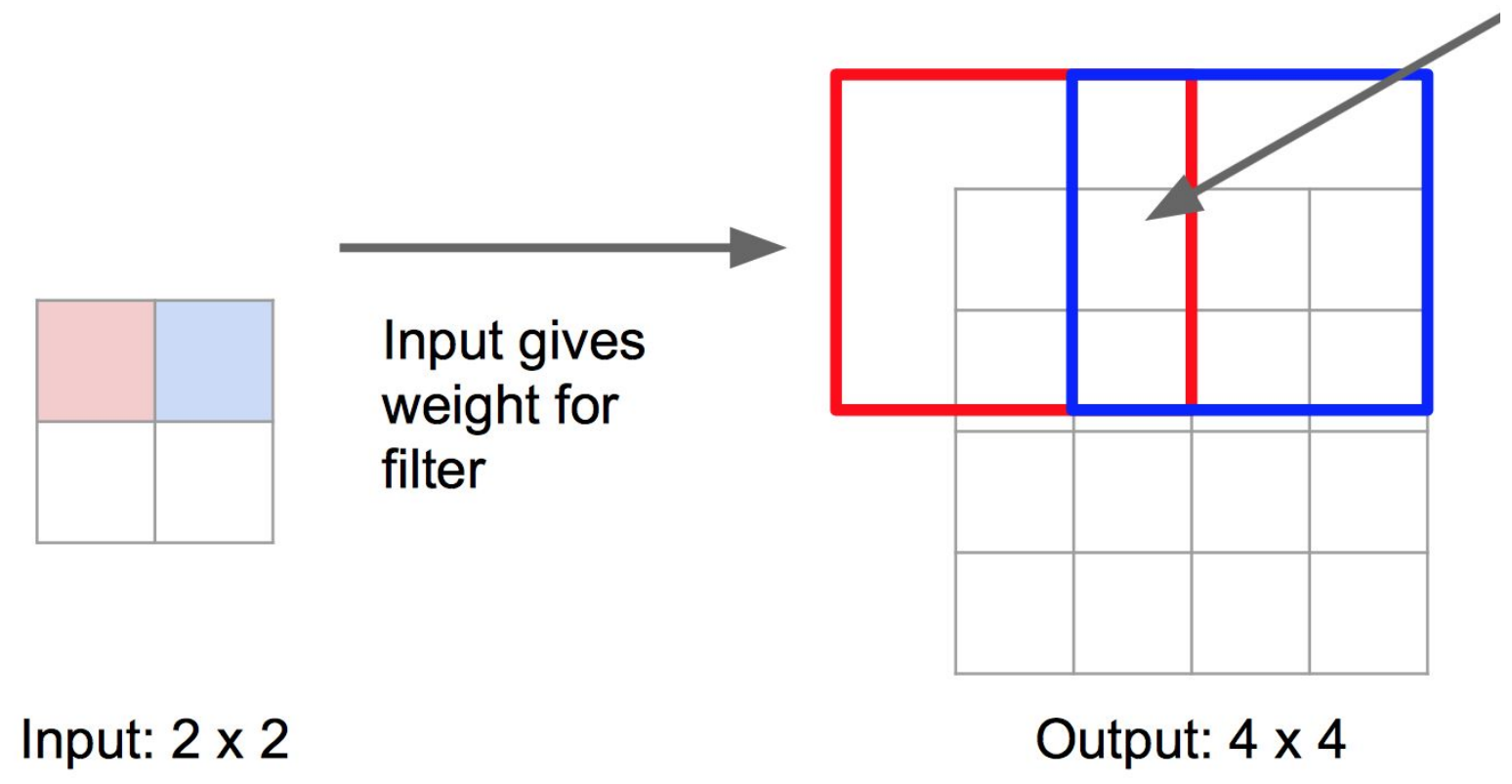

*Figure 1:* Visualization of transpose convolution (Li, et al., 2018)

To visualize the transpose convolutional model, first consider an input image with dimensions 2x2 that is upsampled to dimensions 4x4. To do this, one multiplies the value of a pixel in the downsampled image (left red) with the weights of a filter (right red). The resulting values are then placed in the output matrix, where the filter provides a location for where the values go. The filter will make a stride, which means it will shift a certain amount of units. Strides are calculated from the ratio between the side dimensions of the output and input images. The ratio of side dimensions of the output to the input is 2 to 1. The filter (right blue) then takes a stride of 2 on the output matrix, and that corresponds to the adjacent (left blue) pixel on the smaller image. If there is an overlap in the output matrices, the values are added together. In this case, those would be the column that the arrow on the right points to in Figure 1.

## 2. Land cover change detection via semantic segmentation

### 2.1 Semantic segmentation method

An alternative proposition for a semantic segmentation model architecture is the use of an atrous convolution when downsampling, which is intended to provide dense feature extraction and field-of-view enlargement. An algorithm that uses this is the Deeplab v3+ semantic segmentation model, which is open sourced by Google (Figure 2)

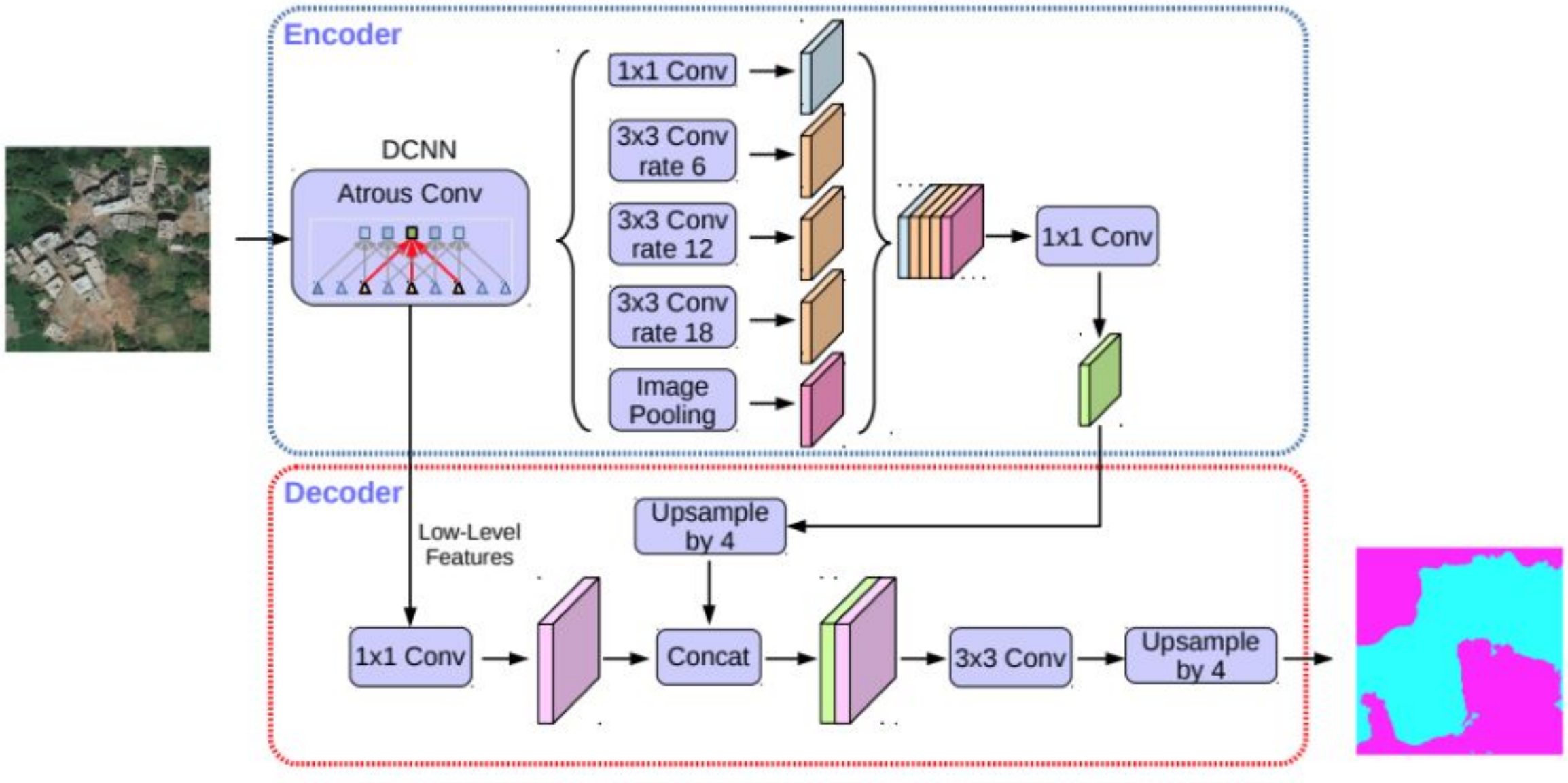

*Figure 2:* Deeplab v3+ architecture (Chen, et al., 2018)

Atrous convolutions upsample the filters themselves, rather than the image resolutions. In the hierarchy of CNN layers, atrous convolution filters are applied after downsampling a few times. An output stride is the number of pixels of the original image represented by a resulting pixel of the new map. For example, in a normal CNN, one would downsample the image resolution so that at the end, each pixel of the response map represents 256 pixels of the original image, which means the output stride is 256. With atrous convolutions, one could stop downsampling at an output stride of 16, with atrous convolution filters being applied subsequently. This filter is similar to the CNN filters, in that it is a matrix of weights, but it is essentially dilated.

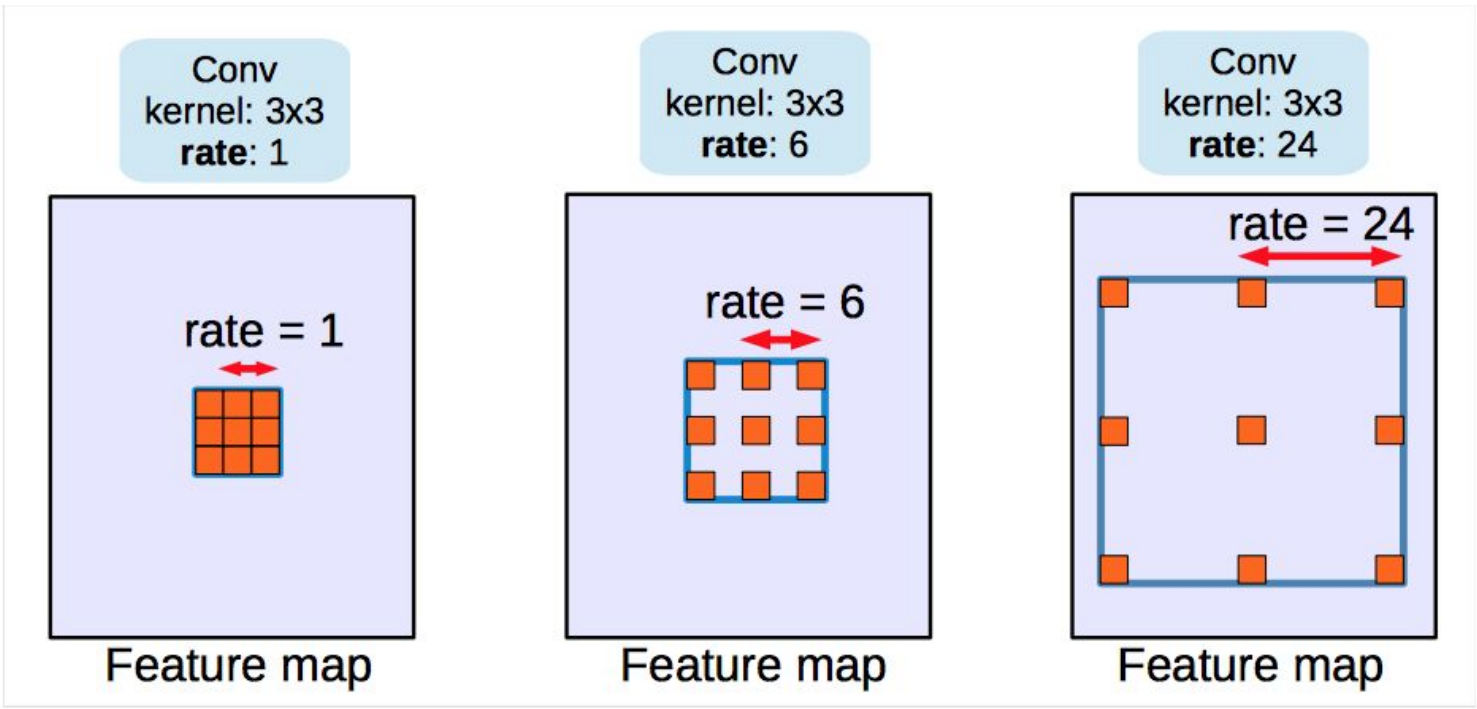

*Figure 3.* Visualization of atrous convolution (Chen, et al., 2017)

The dilation comes from placing zeros between values in the filter, which is shown in Figure 3 (Chen, et al., 2017). Atrous convolution filters have dilation rates, which can be thought of as the distance, or number of pixels, between two adjacent weights. For example, a dilation rate of 1 would result in 0 spaces, a rate of 2 would result in 1 space, and a rate of 6 would result in 5 spaces between weights. This type of convolution is advantageous because it does not require extra parameters in each filter, but simultaneously expands the field of view for the filters (Chen, et al., 2017). Since one of the goals of semantic segmentation is allow for a broader scene understanding, the implementation of atrous convolutions would be incredibly useful.

When choosing dilation rates, there are some cases in which the dilated filter acts as a 1x1 convolution. This is evident when the rates are especially large compared to the dimensions of the original image. For example, if a filter is so large that only the center value of the matrix actually occupies space over the image, then the other values of the filter will be useless, because they cannot be multiplied with actual pixels. One way to deal with this is through a technique called Atrous Spatial Pyramid Pooling (Chen, et al., 2017). This helps with processing objects at multiple scales. After multiple atrous convolutions are applied, atrous spatial pyramid pooling (ASPP) occurs. ASPP applies convolutional feature layers with filters at multiple dilation rates and effective field of views, allowing objects and image context to be captured at multiple scales. In addition, ASPP incorporates image-level features using global average pooling (taking the average of the pixel weights of each feature map) to add more global context information.

Another effective component used in the Deeplab v3+ model is the decoder module. It concatenates a lower level atrous convolution feature map with a higher level feature map, which refines segmentation results along object boundaries.

**2.2 Training pipeline**

The whole process was divided into two components: the training module, and the land cover change detection. The training component is visualized in Figure 4. With our land cover dataset, we started the training data preparation (Figure 6). Then, we used the DeepLab model to train our land cover semantic segmentation model. Finally, we exported the trained model to an inference graph, which was used for land cover prediction followed by land cover change detection.

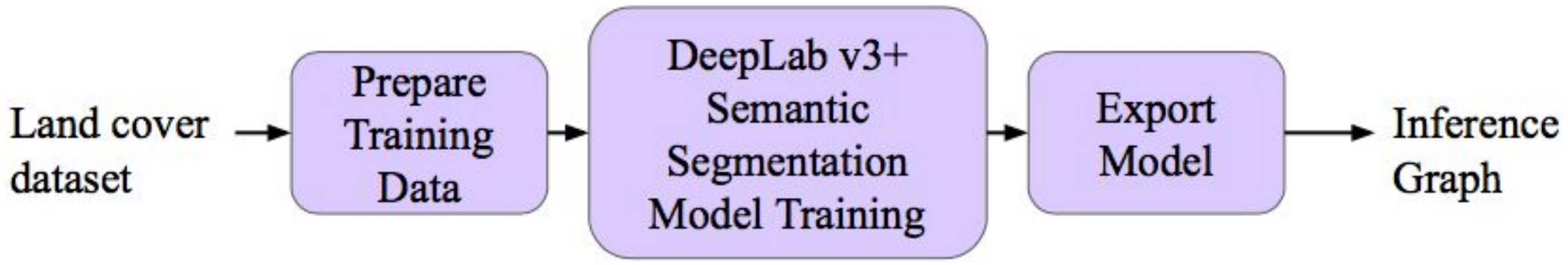

*Figure 4*: Training Pipeline

The training data, which consisted of aerial imagery, were accessed from https://competitions.codalab.org/competitions, under the Land Classification Challenge. There were 803 RGB training images, each of size 2448 by 2448 pixels, with ground sample distance of 50 cm per pixel. Each training image was also paired with a ground truth segmentation mask image, which was used for land cover annotation (Figure 5). Given the training data, the program had seven land cover types to consider: barren land, rangeland, forest, water, urban, agricultural, and unknown. In order to measure the accuracy of the semantic segmentation model, the Intersection over Union (IoU) metric was used. This metric measured the area of overlap between a class label in the ground truth segmentation map and the area of the same class label in the predicted segmentation mask, divided by the area of the union between the two. For this model, the average of the IoUs from each of the seven land cover classes was measured (also called the mean IoU, or mIoU).

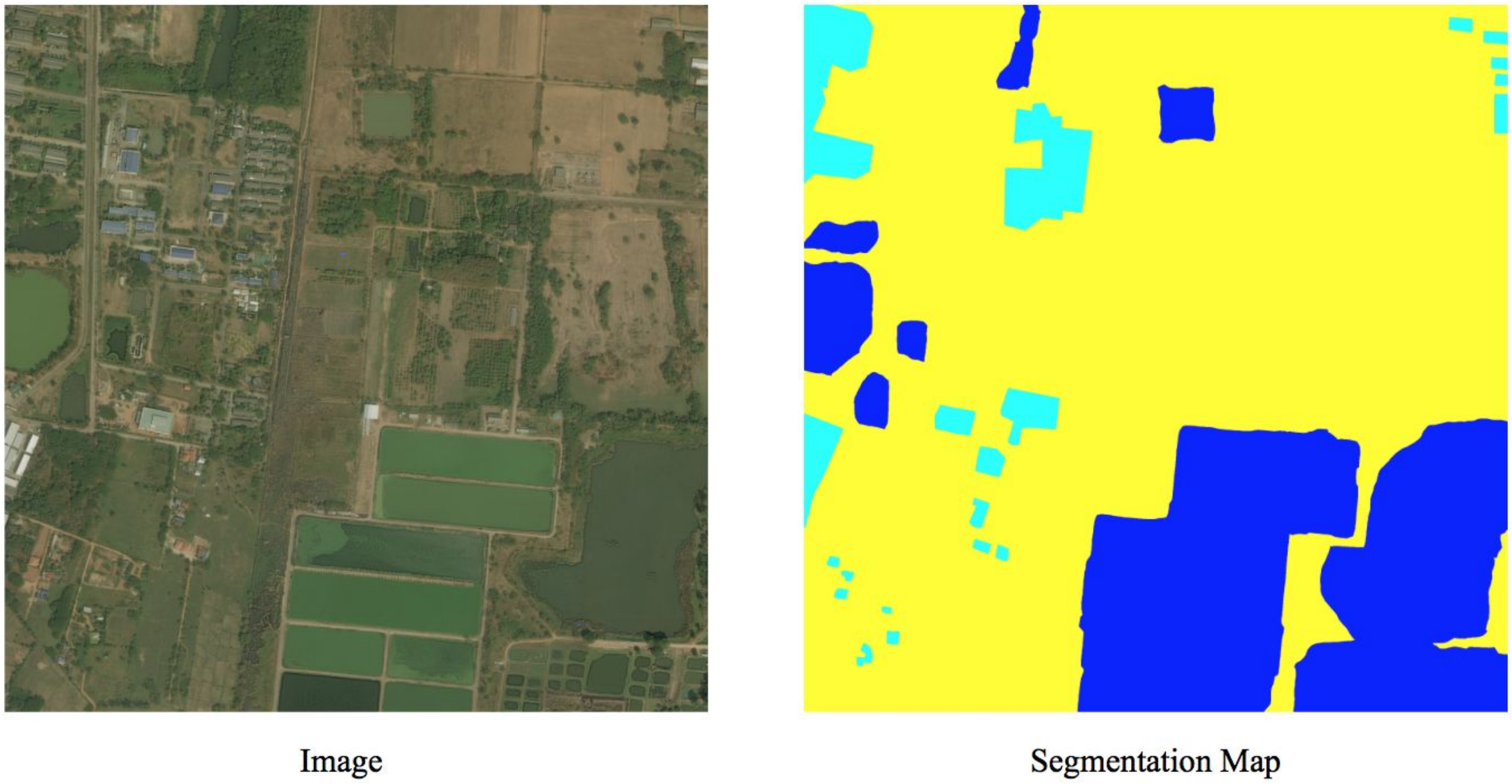

7 land cover types:

- **Urban land**
- **Agriculture land**
- **Rangeland**
- **Forest land**
- **Water**
- **Barren Land**
- **Unknown**

*Figure 5:* Land cover dataset example with color notation for each land cover type

For training data preparation (Figure 6), we paired the images with their corresponding segmentation masks, and proceeded to randomly split the images into two groups, to either training or evaluation. 90 percent of the data were trained, and the remaining 10 percent were used for evaluation. Since the images in the dataset were originally 2448 by 2448 pixels, we divided each image (and its corresponding segmentation mask) into tiles of 512 by 512 pixels, resulting in 25 images of a much smaller size to be used for training. Afterwards, we converted each pair into the Tensorflow example proto, which were subsequently written to a TFRecord file.

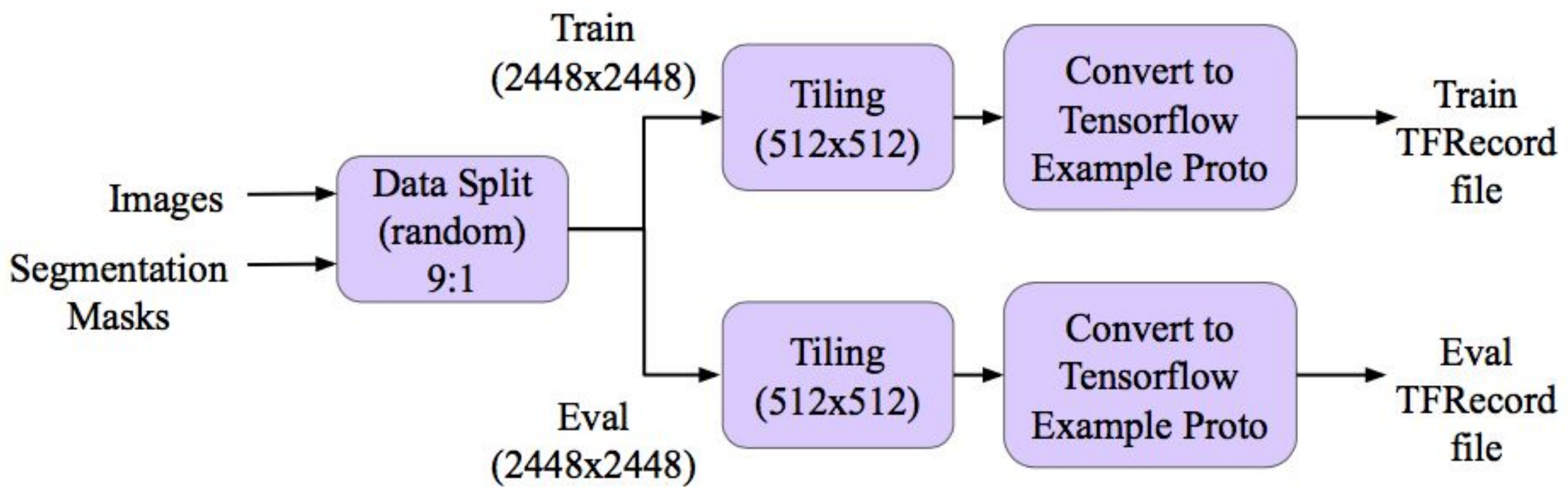

*Figure 6*: Training Preparation

**2.3 Land cover change detection**

After the model had been trained, it was utilized for change detection (Figure 7). For the inference, we studied forest land, grass land and urban land in the United States, through a time difference of at least ten years. They images were mainly of sites that had undergone urban development or deforestation in the 21st century. We used satellite imagery from Star, Idaho; Orlando, Florida; and Morgan Hill, California. Using the Historical Image feature on Google Earth, we retrieved images from two different years ($T_1$and $T_2$) at those locations. We then cropped the images to 512 by 512 pixels, and applied the inference graph to predict the segmentation map. Using the segmentation map, we calculated the land cover percentage for each class type present in each image.

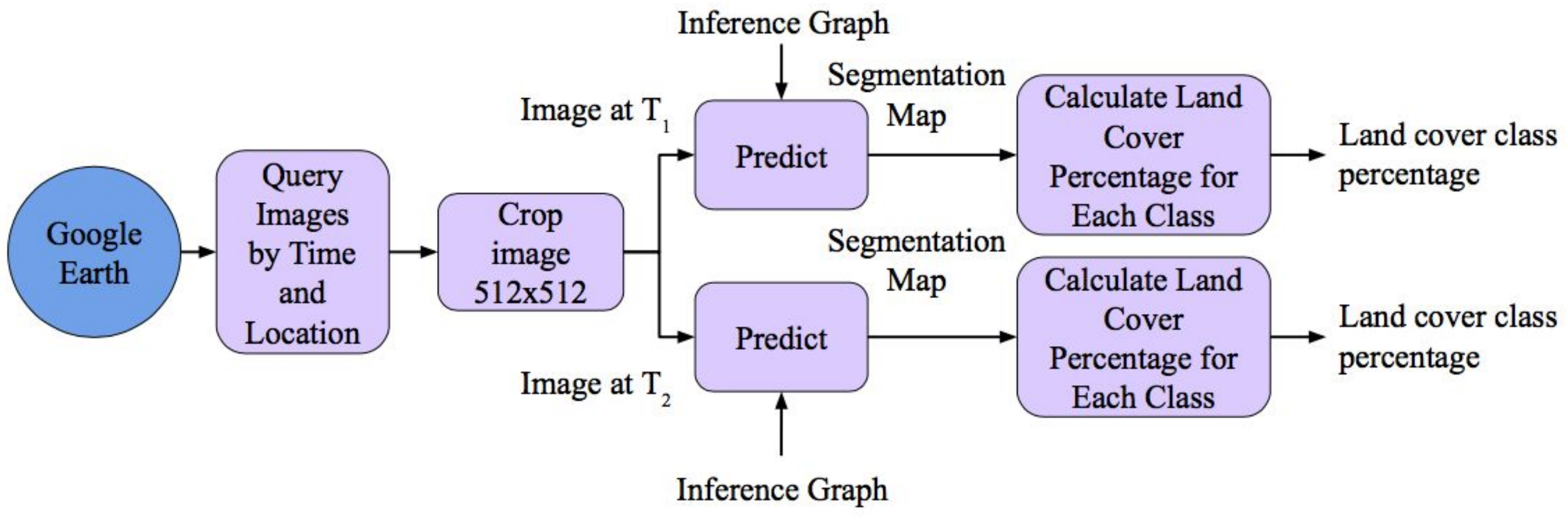

*Figure 7*: Land cover change detection

## 3. Results

From training the model, we achieved a mean IoU of 75.6% on our evaluation set. This was significantly higher than the mean IoU of 43.3% from the DeepGlobe challenge (Figure 8). This difference was due to the fact that the model from the challenge used Resnet18 as its network backbone, while our model used the improved Xception backbone. Furthermore, the newer version of Deeplab used atrous convolutions, which effectively enlarged the field of view and allowed for dense feature extraction. Finally, a decoder module was included, which refined the segmentation results, especially along object boundaries.

| | mIoU |
|---|---|
| DeepGlobe land cover classification challenge (Chen, et al., 2018) | 43.3% |
| **Ours** | **75.6%** |

*Figure 8*: Performance comparison

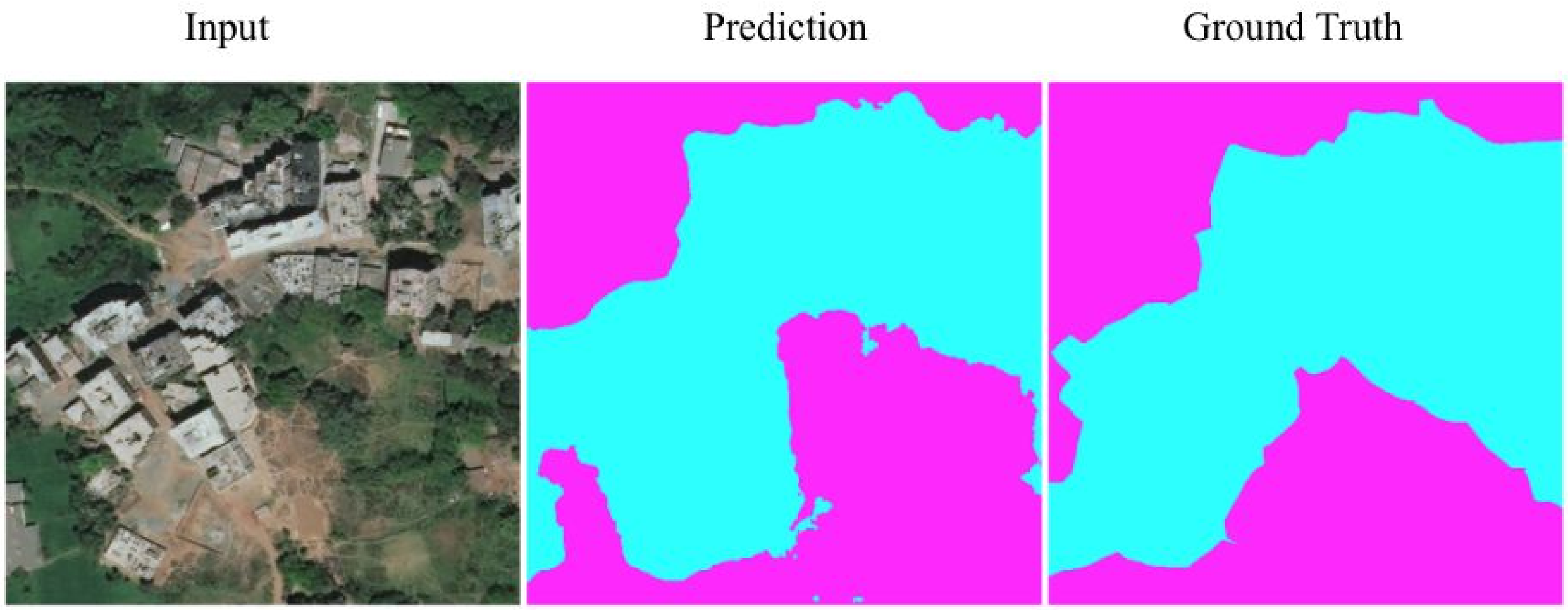

*Figure 9*: Evaluation example

In the evaluation set, it was noted that the model could categorize land cover types beyond the scope of an actual person. Figure 9 illustrates a side-by-side comparison of the prediction from the trained model and ground truth masks of an example input image from the evaluation set. In the lower left corner of the input image, there was a stretch of green land that was incorrectly labeled as urban land in the ground truth and was predicted correctly as rangeland in the inference. In this case, the model was more accurate than the person who labeled the ground truth, because it was able to recognize the vegetation in the area.

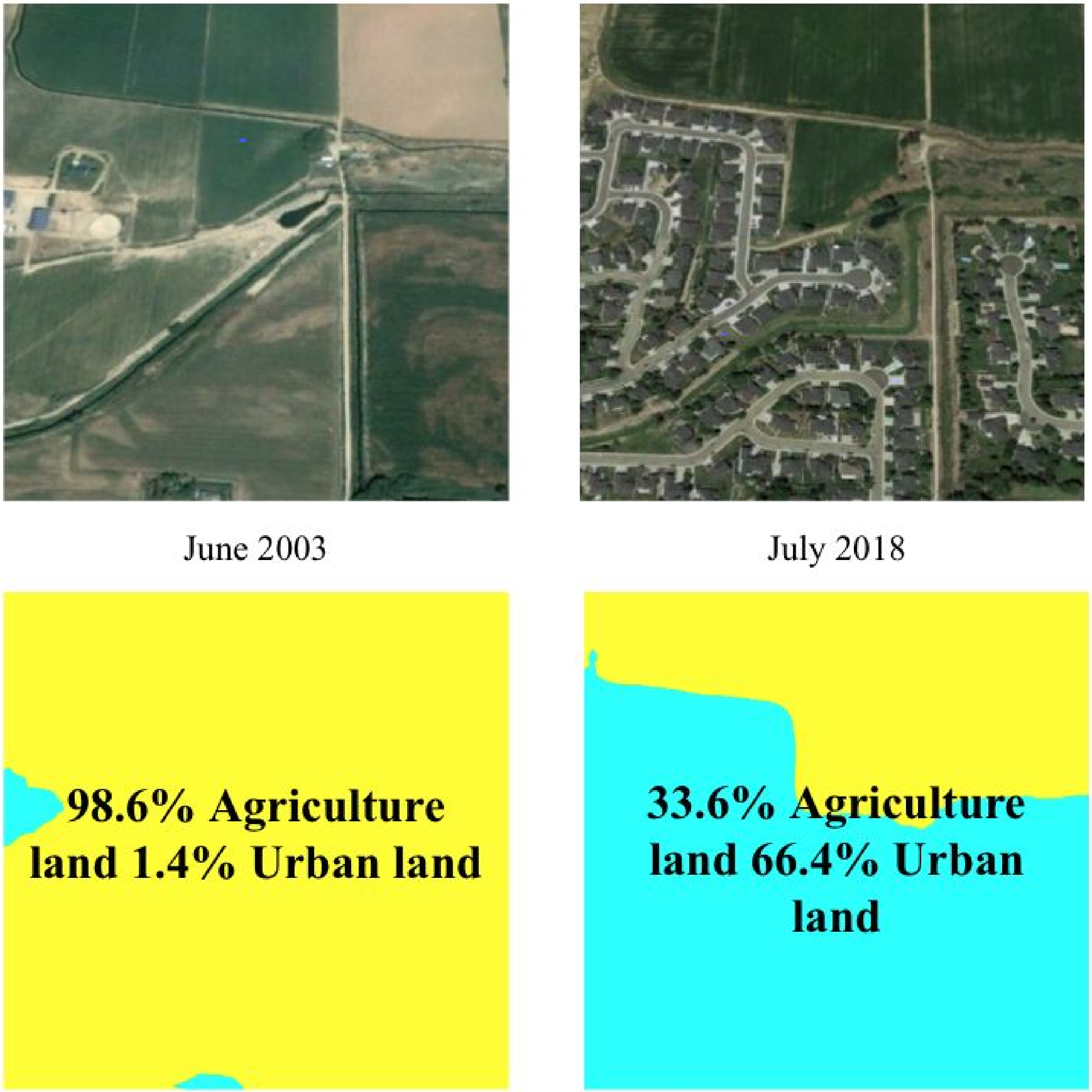

*Figure 10:* Star, Idaho, over a fifteen year period

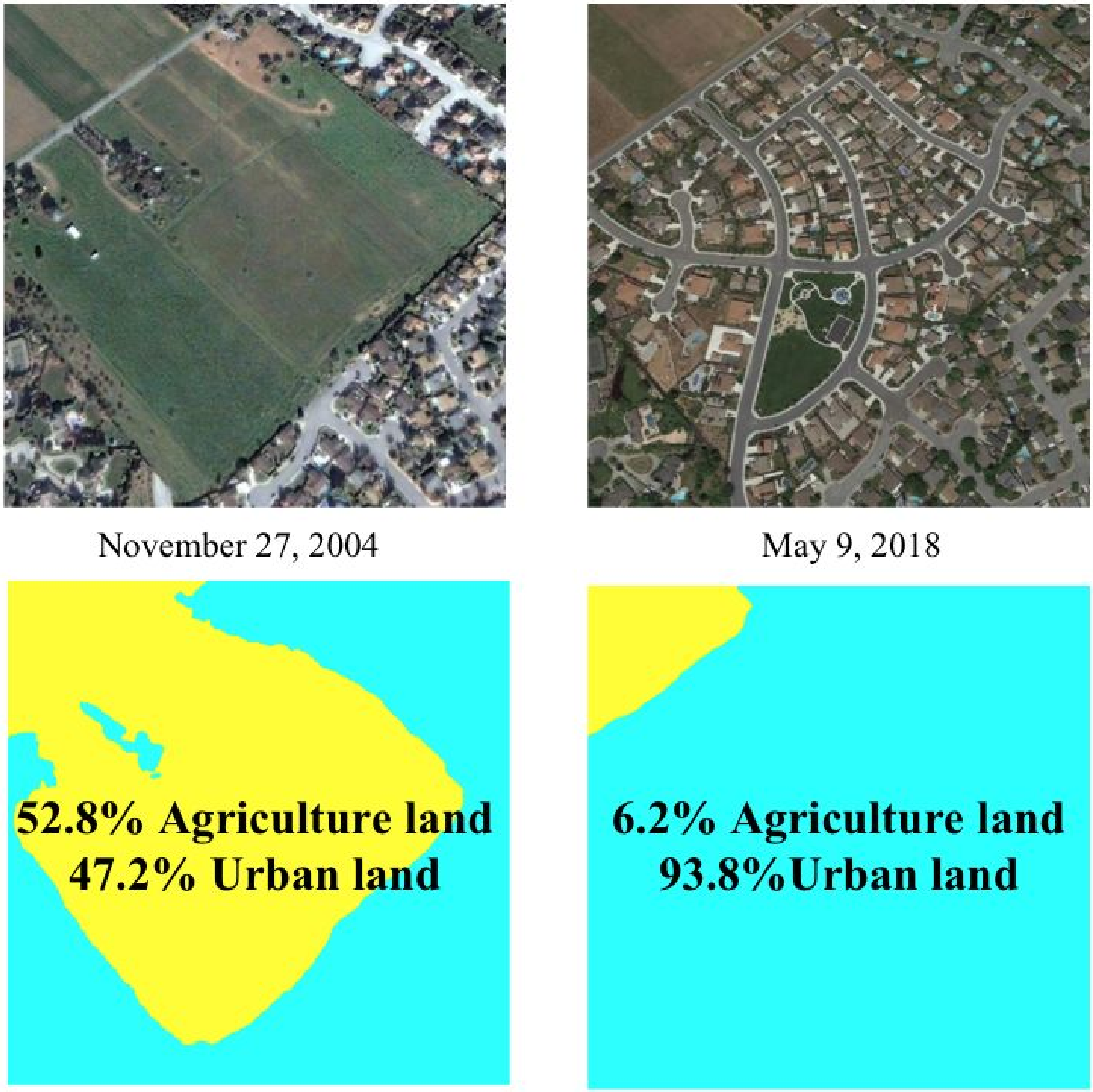

*Figure 11:* Morgan Hill, California, over a fourteen year period

Figures 10 and 11 were two locations retrieved from Google Earth that we applied change detection on. In the area of Morgan Hill, California, one can see that in 2004, agriculture and urban land were each occupied a roughly equal percentage of the land. However, in 2018, our model predicted that urban land percentage had almost doubled, leaving only 6.2 percent of the land in the image as agriculture land. Similarly, in the area of Star, Idaho, 2003, most of the land was for agriculture. In 2018, about two-thirds of that land had been built into a residential area. The model prediction in the second location roughly outlines the boundaries of the neighborhood shown. Below is a summary of three locations that the model performed on (Figure 12). For all three places, the amount of urban land had a significant increase.

| | Images | | | | | |
|---|---|---|---|---|---|---|
| | Morgan Hill area, California | | Star area, Idaho | | Orlando area, Florida | |
| | Time Period | | Time Period | | Time Period | |
| Land Cover Type | November 2004 | May 2018 | June 2003 | July 2018 | December 2004 | November 2011 |
| Forest | 0% | 0% | 0% | 0% | 100% | 0% |
| Agriculture | 52.8% | 6.2% | 98.6% | 33.6% | 0% | 0% |
| Urban | 47.2% | 93.8% | 1.4% | 66.4% | 0% | 100% |

*Figure 12:* Data table summarizing percentage of land cover types

## 4. Conclusions

We achieved significant improvement in land cover semantic segmentation, as we obtained a mIOU of 0.756 compared to the value 0.433, which was reported in the DeepGlobe land cover classification challenge. We also proposed the land cover change detection system that leveraged the state-of-the-art method of semantic segmentation. Applications of this model include land management, in regard to natural resources, deforestation analysis, and changes in the environment.

There are future works that would be further improve the system. For model generalization, we should add data augmentation to simulate images from Google Earth without proper color correction or haze removal. For change detection in a large satellite image, we would need to divide the image into overlapped tiles, and then apply inference on each tile. Then we would study how to merge the segmentation results in those overlapped areas.

For future applications, it would be useful to apply this model to places undergoing rapid land cover changes, such as timber industries and real estate development. In addition, one caveat of our change detection system is that none of the images retrieved from Google Earth were paired with ground truth labels. To further improve the model, we would provide ground truth masks for Google Earth images on which we would measure change detection.